\documentclass{ecai} 

\usepackage{latexsym}
\usepackage{amssymb}
\usepackage{amsmath}
\usepackage{amsthm}
\usepackage{booktabs}
\usepackage{enumitem}
\usepackage{graphicx}
\usepackage{color}

\usepackage{microtype}
\usepackage{booktabs} 

\usepackage{tabularx}
\usepackage{multirow}
\usepackage{hyperref}

\usepackage{amsfonts}
\usepackage{amssymb}
\usepackage{mathtools}
\usepackage{amsthm}
\usepackage{soul}
\RequirePackage{algorithm}
\RequirePackage{algorithmic}
\usepackage{caption}
\usepackage{subcaption}
\usepackage{xcolor}

\theoremstyle{plain}
\newtheorem{theorem}{Theorem}
\newtheorem{lemma}{Lemma}

\theoremstyle{definition}

\theoremstyle{remark}
\newtheorem{remark}[theorem]{Remark}

\newcommand{\BibTeX}{B\kern-.05em{\sc i\kern-.025em b}\kern-.08em\TeX}

\DeclareMathOperator*{\argmax}{argmax}

\begin{document}


\begin{frontmatter}


\paperid{364} 


\title{General Lipschitz: Certified Robustness Against Resolvable Semantic Transformations via Transformation-Dependent Randomized Smoothing}




\author[A,B]{\fnms{Dmitrii}~\snm{Korzh}
\thanks{Corresponding Author. Email: korzh@airi.net.}
}
\author[B, A,C]{\fnms{Mikhail}~\snm{Pautov}
} 
\author[D, E]{\fnms{Olga}~\snm{Tsymboi}
}

\author[B, A]{\fnms{Ivan}~\snm{Oseledets}
}

\address[A]{Skolkovo Institute of Science and Technology, Moscow, Russia }
\address[B]{Artificial Intelligence Research Institute, Moscow, Russia}
\address[C]{
ISP RAS Research Center for Trusted Artificial Intelligence, Moscow, Russia}
\address[D]{Moscow Institute of Physics and Technology, Moscow, Russia}
\address[E]{Sber AI Lab, Moscow, Russia}


\begin{abstract}
Randomized smoothing is the state-of-the-art approach to constructing image classifiers that are provably robust against additive adversarial perturbations of bounded magnitude. However, it is more complicated to compute reasonable certificates against semantic transformations (e.g., image blurring, translation, gamma correction) and their compositions. In this work, we propose General Lipschitz (GL), a new flexible framework to certify neural networks against resolvable semantic transformations.  Within the framework, we analyze transformation-dependent Lipschitz-continuity of smoothed classifiers w.r.t. transformation parameters and derive corresponding robustness certificates. To assess the effectiveness of the proposed approach, we evaluate it on different image classification datasets against several state-of-the-art certification methods. 
\end{abstract}

\end{frontmatter}

\section{Introduction}
\label{introduction}
Deep neural networks show remarkable performance in a variety of computer vision tasks. However, they are drastically vulnerable to specific input perturbations (called adversarial attacks) that might be 
imperceptible
to the human eye as it was initially shown in \cite{szegedy2013intriguing, biggio2013evasion}.  
Namely, suppose that deep neural network $f: \mathbb{R}^n \to [0,1]^C$ maps input images $x$ to class probabilities. Then, given the classification rule $\hat f(x) = \arg\max_{i \in Y} f_{i}(x)$, where $Y = \{1, 2, \dots, C\}$,
it is possible to craft an adversarial perturbation $\delta$ of small magnitude such that $x$ and $x + \delta$ are assigned by $\hat{f}$ to different classes. 
For some applications, such as self-driving cars 
\cite{tu2021exploring} and identification systems 
\cite{komkov2021advhat,pautov2019adversarial}, this vulnerability to the small change in the input data is a serious concern.


\begin{figure*}[ht]
  \centering
   \includegraphics[width=.7\textwidth]{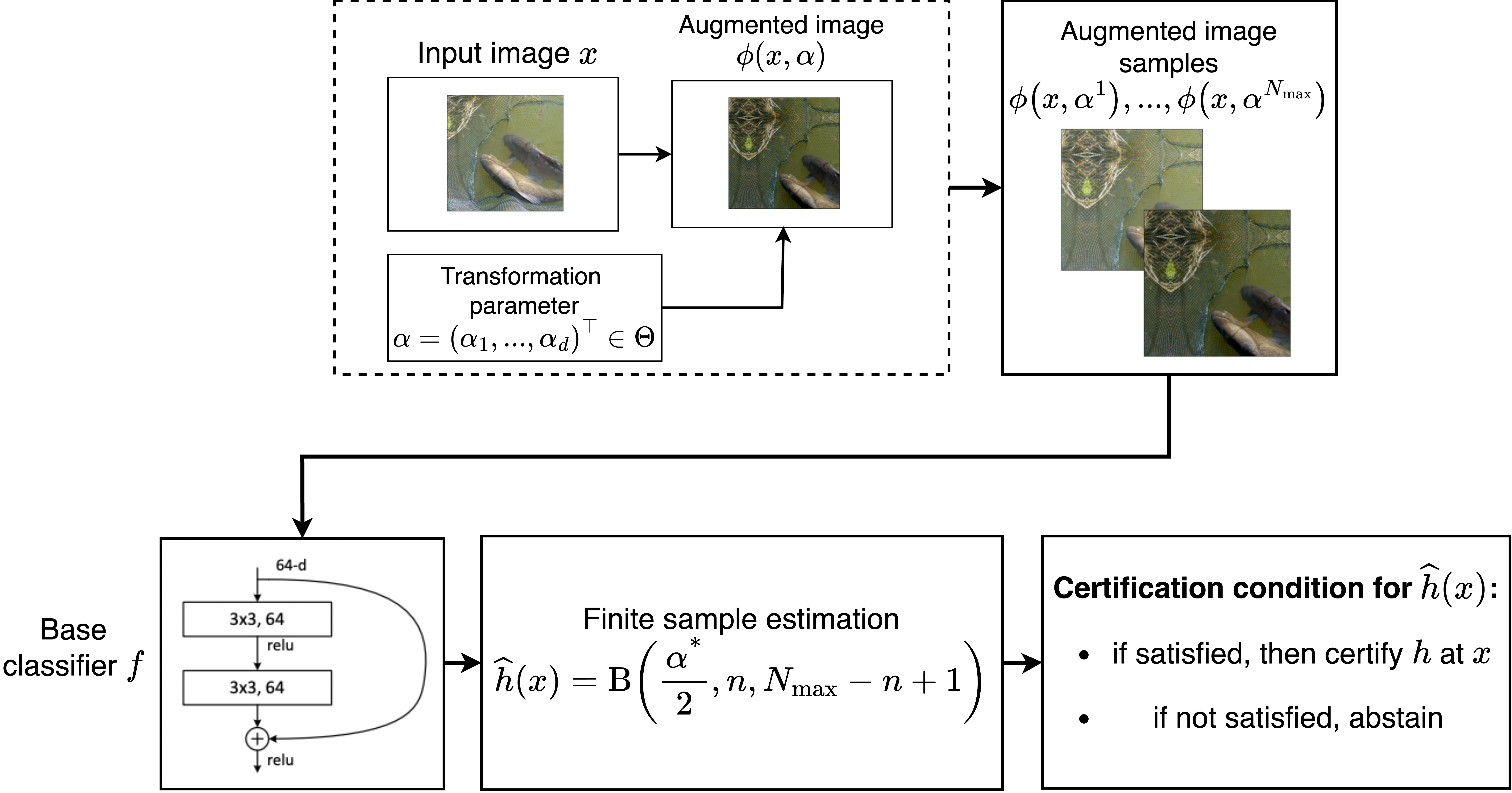}
   \vspace{1em}
  \caption{Schematic illustration of the certification procedure: given an input image $x$ of class $c$ and parametric transformation $\phi$, we sample $N_\text{max}$ augmented images $\{\phi(x, \alpha^1), \dots, \phi(x, \alpha^{N_\text{max}})\}$ and compute a lower bound on $h_c(x)$ using Clopper-Pearson test in form $\hat{h}_c(x) = B(\alpha^*/2, n, N_\text{max} -n +1),$ where $B$ is Beta distribution and $n$ is the number of augmented images $\phi(x, \alpha^j)$ for which the value of the base classifier $f_c(\phi(x, \alpha^j)) > \frac{1}{2}$. Then, the certification condition from Theorem \eqref{theor:main} is checked for the value $\hat{h}_c(x).$}
  \label{fig:teaser}
  \vspace{1em}
\end{figure*}

Recently, many approaches to create adversarial perturbations were proposed, as well as defense techniques to counteract these approaches, causing an attack-defense arms race \cite{akhtar2018threat,xu2020adversarial}. 
This race, however, barely affected neural network applications where the provably correct behavior of models is required.  
As a result, more research was conducted in the field of certified robustness,  where the goal is to provide provable guarantees on the models' behavior under different input transformations. Randomized smoothing  \cite{lecuyer2019certified,cohen2019certified} is among the most effective and popular approaches used to build provably robust models. 
Initially developed as a certification tool against norm-bounded additive perturbations, it was later extended to the cases of semantic perturbations \cite{li2021tss,hao2022gsmooth}, such as brightness shift and translations.
In the case of additive perturbations, this approach is about replacing the base classifier $f$ with a smoothed one in the following form:
\begin{equation}\label{gauss:rs}
    g(x) = \mathbb{E}_{\varepsilon \sim \mathcal{N}(0, \sigma^2)} f(x + \varepsilon).
\end{equation}
As it is shown in prior works \cite{lecuyer2019certified, cohen2019certified}, if the base classifier predicts well under Gaussian noise applied to input $x$, then the smoothed one is guaranteed not to change the predicted class label
in some vicinity of $x$. Lately, it was shown \cite{li2021tss,hao2022gsmooth} that if the input is subjected to semantic transformation with the parameters sampled from a particular distribution, then it is possible to derive the robustness guarantees for a smoothed model against corresponding semantic transformation.

In this work, we focus on resolvable \cite{li2021tss} semantic transformations and their compositions and develop a new framework for certified robustness of classifiers under these perturbations. 
In a nutshell, we approach smoothing from a different angle and show that a smoothed classifier is Lipschitz continuous with respect to parameters of compositions of resolvable transformations. 

Our method is derived with no assumptions on semantic transformation and smoothing distribution (except for the resolvability and smoothness, respectively). It provides a constructive numerical procedure for building a certification against a particular transformation. The proposed approach scales to large datasets at the cost of inference of the smoothed model and can be applied for certification against compositions of resolvable semantic transformations. 

\textbf{Our contributions are summarized as follows:}
\begin{itemize}
    \item We propose a universal certification approach against compositions of resolvable transformations based on randomized smoothing. Our method can be applied for certification against any composition of resolvable semantic transformations, in contrast to the previous studies. 
    \item We propose a numerical procedure to verify the smoothed model's robustness with little to no computation overhead.
    \item We evaluate our method on different datasets and show that it yields state-of-the-art robustness guarantees in the majority of considered experimental setups. 
\end{itemize}

\section{Preliminaries}

In this work, we focus on the task of image classification. 
Given $X \subset \mathbb{R}^n$ as the set of input objects and   $\mathcal{Y} = \{1, 2, \dots C\}$ as the set of classes, the goal is to construct a mapping $\hat f: X \to \mathcal{Y}$ assigning a label to each input object. 
Following the convenient notation, this mapping may be represented as 
\begin{equation}
\hat f(x) = \argmax_{i \in \mathcal{Y}} f_{i}(x),
\end{equation}
where $f: X \to [0, 1]^C$ is a classification model and $f_{i}(x)$ refers to $i-$~th predicted component. 

Suppose a parametric mapping $\phi: X \times \Theta \to X$ corresponds to a semantic perturbation of the input of the classification model, where $\Theta$ is the space of parameters of the perturbation. 
The goal of this paper is to construct the framework to certify that a classifier
is robust at $x \in X$ to the transformation $\phi(x, \cdot)$ for some set of parameters $\mathcal{B}(\beta_{0})$, where $\phi(x, \beta_{0}) =  x$ for all 
$x \in X$.

A transform $\phi: X \times \Theta \to X$ is called \emph{resolvable} \cite{li2021tss} if for any parameter $\alpha~\in~\Theta$ there exists a continuously differentiable function $\gamma: \Theta \times \Theta \to \Theta$ such that for all $x \in X$ and all $\beta \in \Theta$
\begin{equation}
\label{eq:resolvable_transform}
 \phi(\phi(x, \alpha), \beta) = \phi(x, \gamma(\alpha, \beta)).
\end{equation}
In this work, we analyze the Lipschitz properties of randomized smoothing to certify classification models against compositions of resolvable transformations.  
\section{Proposed method}

This section is devoted to the proposed certification approach and its theoretical analysis.

\subsection{Randomized smoothing for semantic transformations}
For the given base model $f: X \subset \mathbb{R}^n \to [0,1]^C$, input image $x \in X$, resolvable transformation $\phi: X \times \Theta \to X$ with the resolving function $\gamma$ as defined in \ref{eq:resolvable_transform}, we construct the smoothed classifier $h(x)$ in the form of expectation over perturbation density $\rho(y \vert x)$ conditioned on the observed sample $x$:
\begin{align}\label{r:smoothing}
    \begin{split}
    h(x) &= \int_{\Theta} f(\phi (x, \alpha)) \rho(\phi (x, \alpha) \vert x) d \alpha \\ &= \int_{\mathbb{R}^n} f(y) \rho(y \vert x) dy.
    \end{split}
\end{align}
The goal of this paper is to present a procedure that guarantees a smoothed model to be robust to semantic perturbations, that is
\begin{equation} \label{eq:robustness}
    \argmax_{i \in \{1, 2, \dots C\}} h_{i}(x) = \argmax_{i \in \{1, 2, \dots C\}} h_{i}(\phi(x, \beta)),
\end{equation}
for all $\beta \in \mathcal{B}(\beta_0)$, where $\phi(x, \beta_0) = x.$
One way to achieve robustness to parametric perturbation is to bound the Lipschitz constant of the classifier from Eq. \eqref{r:smoothing} with respect to the transformation parameters. 
For this purpose, the density $\rho(y| x)$ has to be continuously differentiable with respect to perturbation parameters; otherwise, this problem becomes ill-posed. To overcome this issue, we introduce additional Gaussian smoothing.

Assuming that the perturbed sample has the form $\hat{x} = \phi(x, \beta)$, we redefine an auxiliary variable $y = \phi(\hat{x}, \alpha) + \varepsilon$, where $\varepsilon \sim \mathcal{N}(0, \sigma^2 I_n)$ and the overall conditional probability density $\rho(y|\hat{x})$ in the form: 
\begin{equation}
    \label{eq:density}
    \rho(y|\hat{x}) = \frac{\int_{\Theta} \exp\left\{-\frac{\|y-\phi(\hat{x}, \alpha)\|_2^2}{2\sigma^2}\right\} \tau(\alpha) d\alpha}{(2\pi\sigma^2)^{\frac{n}{2}}},
\end{equation}
where $\tau(\alpha)$ is the smoothing distribution of the transformation. Following the literature, \cite{hao2022gsmooth, li2021tss}, we sample $\alpha \sim \mathcal{N}(0, \sigma_{\alpha}^2 I_d)$, and then map it to the desired smoothing distribution (see Section \ref{s:experiments} for details). Here, $d = \mathrm{dim}(\Theta)$ is the number of transformation parameters.

\subsection{Robustness guarantee}

In this section,  we discuss the main theoretical result. 
Prior works mainly concentrate on estimating global Lipschitz constants, which may lead to loose guarantees.
Instead, we provide certification conditions based on local properties due to using perturbation-dependent smoothing.

Let $x \in X$ be the input object of class $c$ and assume 
that the smoothed classifier $h$ defined in Eq. \eqref{r:smoothing} correctly classifies $x$ with significant confidence, i.e., $h_c(x) > \frac{1}{2}.$ Then, the following result holds.

\begin{theorem}{Certification condition.}\label{theor:main}

\noindent
Let $\beta(t): [0,1] \to \Theta$ be a smooth curve such that $\beta(0) = \beta_0$ and $\beta(1) = \beta$. Then there exist mappings $\xi: \left[0, 1\right] \to \mathbb{R}$ and  $\hat g(\beta): \Theta \to \mathbb{R}$ such that if 
$\hat g(\beta) < - \xi(1 - h_c(x)) + \xi(1/2) $, then $h$ is robust at $x$ for all $\beta \in \beta(t),$ where $t \in [0, 1]$.

\begin{proof} (Sketch)

Let $x$ be a fixed input of class $c$. Reassign $h(\beta) = h_c(\beta)$ to manipulate only with $c-$th component of the smoothed classifier that confidently and correctly classifies the ground truth class, namely, let $h_c(\beta) > \frac{1}{2}.$  

To construct the certification criteria, we observe that a directional derivative of $h(\beta)$ with respect to $\beta$ is bounded by the product of two functions, namely $p: [0,1] \to \mathbb{R}$ and $g: \Theta \to \mathbb{R}_{\ge 0}$ such that 
\begin{equation}
   \langle \nabla_\beta\  h(\beta), u \rangle = \int_{\mathbb{R}^n} f(y) \langle \nabla_\beta \rho(y|\hat{x}),u \rangle dy \le p(h(\beta)) g(\beta) 
\end{equation}
for all $u: \|u\|_2=1$. Note that such $p(\cdot)$ and $g(\cdot)$ exist since $h(\cdot)$ is assumed to be smooth (e.g., $p \equiv 1, g(\beta) \equiv \sup_{u} \sup_{\beta} \langle \nabla_{\beta} h(\beta), u\rangle$). 
\begin{equation}\label{eq:nabla_beta_proof_0}
    \langle\nabla_\beta h(\beta), u \rangle = \int_{\mathbb{R}^n} f(y) \eta(y, \hat{x}) \rho(y|\hat{x}) dy,
\end{equation}
where $\eta(y, \hat{x}) = \langle \nabla_\beta \log \rho(y|\hat{x}), u \rangle$ and $u$ is fixed.
To estimate the $\tilde g(h, \beta) \leq p(h)g(\beta)$ supremum in Eq. \eqref{eq:nabla_beta_proof_0} given $\beta$, we solve an optimization problem with a constraint on current fixed value of $h$, that change limits of integration.
Then, integrating this inequality along a smooth curve $\beta(t): \beta(0) = \beta_0, \beta(1)=\beta$, we get 
\begin{equation}\label{eq:how_xi_0}
    \int_{\beta(t)} \langle \nabla_\beta h(\beta), u\rangle dt \le \int_{\beta(t)} p(h) g(\beta) dt.
\end{equation}
Introducing an auxiliary function
\begin{equation}
    \xi(h) = \int \frac{1}{p(h)}dh
\end{equation}
we get 
\begin{equation}\label{eq:bound_fin_0}
\begin{aligned}
    & \xi(h(\beta)) - \xi(h(\beta_0)) = \int_{\beta(t)} \langle \nabla_{\beta} \xi(h(\beta)), u\rangle dt
    \\
    & \le \int_{\beta(t)}g(\beta)dt = \hat{g}(\beta).
\end{aligned}
\end{equation}
Note that $\xi$ is a monotonically increasing function w.r.t. $h(\beta)$ according to the definition (i.e. $\rho(h) \ge 0$); and $\hat{g}(\beta)$ is non-decreasing  along $\beta(t)$ since $g(\beta)$ is non-negative. Assuming that there exists $\beta \in \beta(t)$ such that 
\begin{equation}
    h_{\ne c}(\beta) > \frac{1}{2}, 
\end{equation}
where $h_{\ne c}(\beta)$ corresponds to the probability of assigning a sample  \emph{not} to class $c$. Finally, using the monotonous property of $\xi(h)$ yields a contradiction, proving the result.


\end{proof}

\begin{remark}
    The full proof of the theorem is moved to the Appendix so as not to distract the reader. The assumption on $h_c(x) > \frac{1}{2}$ is given to interpret the multiclass classification problem as binary classification (as the one-vs-all setting).
    The procedure of computing the functions $\xi$ and $\hat{g}$ is described in the numerical evaluation section \ref{ss:numerical}. Intuitively, these functions reflect the Lipschitz-continuity of the smoothed classifier w.r.t. transformation parameters. 
\end{remark}
\end{theorem}

The theorem states that the smoothed classifier is robust at the point $x$ to transformation $\phi$ for all parameter values $\beta \in \beta(t),$ if the certification condition is verified for a single parameter value. Note that the certified set of parameters is not necessarily in the ball vicinity of the initial parameter value. This is \emph{the first approach for non-ball-vicinity certification} to our knowledge.

\subsection{Numerical evaluation}
\label{ss:numerical}

The Theorem \eqref{theor:main} anticipates a numerical procedure to compute certification functions $\xi,~\hat{g}$. In this section, we describe this procedure in detail.

Here and below, we assume that the input sample $x$ is fixed and treat smoothed model $h$ as the function of the perturbation parameter $\beta$, namely $h(\phi(x, \beta)) \equiv h(\hat x) \equiv h(x,\beta) \equiv h(\beta)$ for simplicity. Within our framework, functions $\xi,~\hat{g}$ are derived as the ones bounding the smoothed classifier's directional derivative with respect to the perturbation parameter:
\begin{equation}\label{eq:bound_for_dd}
    \langle \nabla_{\beta} h(\beta), \beta  \rangle  \leq \tilde g(h(\beta), \beta) \leq p(h)g(\beta),
\end{equation}
where $\tilde g(h(\beta), \beta)$ is an upper bound on the directional derivative. This function is also bounded by the product of a function of $h$ and the function of $\beta$.
If the functions $p(h)$ and $g(\beta)$ are known, the mappings from Theorem \ref{theor:main} have the following form:
\begin{equation} \label{eq:xi_and_hat_g}
    \xi(h) = \int \frac{1}{p(h)} dh, \quad \hat{g}(\beta) = \int^1_0 g(\beta(t)) dt.
\end{equation}

It is worth mentioning that the function $\xi(h)$ can be derived analytically, for example, for additive transformations with $\tau(\alpha) \sim \mathcal{N}(0, \kappa^2)$, $d=1$:
\begin{equation}
\begin{split}
    \log \rho(y \mid \hat{x})= &-\frac{1}{2} \log(2 \pi) - \frac{1}{2} \log(\kappa^2 + \sigma^2)  -
    \\
    &\frac{(y-x-\beta)^2}{2(\sigma^2 + \kappa^2)}
\end{split}
\end{equation}
\begin{equation}
    \eta(y, \hat x) = \frac{\partial}{\partial \beta} \log \rho(y \mid \hat x) = \frac{y -x -\beta}{\sigma^2 + \kappa^2},
\end{equation}
\begin{equation}
    \tilde g(h, \beta) = \frac{1}{\sqrt{\sigma^2 + \kappa^2}\sqrt{2 \pi}} e^{(\operatorname{erf}^{-1}(1-2h))^2}
\end{equation}
\begin{equation}
    \xi(h) =  \sqrt{\sigma^2 + \kappa^2} \Phi^{-1}(h),
\end{equation}
where $\Phi^{-1}$ is a standard Gaussian distribution's inverse cumulative density function. This result coincides with the one from  \cite{cohen2019certified, li2021tss} if $\sigma =0$.

\subsubsection{Bounding the directional derivative}
 
A bound for the directional derivative in the form from Eq. \eqref{eq:bound_for_dd} is used to compute functions $\xi, \hat{g}$ from Theorem \eqref{theor:main}. However, in the case of a complicated form of conditional density from Eq.  \eqref{eq:density}, it may be unfeasible to construct an exact bound. Instead, we propose to use a numerical procedure to bound directional derivatives of the smoothed model. The gradient of the smoothed classifier with respect to the parameters of transformation has the following form:
\begin{equation}
\begin{aligned}
    \nabla_{\beta} h &= \int f(y) \nabla_{\beta} \rho(y \vert \hat{x}) d y \\
    &=\int f(y) \eta(y, \hat{x}) \rho(y \vert \hat{x}) d y, 
\end{aligned}
\label{eq:nabla_h}
\end{equation}
where $\eta(y, \hat{x})=\nabla_{\beta} \log \rho(y \vert \hat{x})$. Given fixed $\beta,$ the problem of bounding the directional derivative $\langle \nabla_{\beta} h(\beta), \beta  \rangle$ is equivalent to the search of the worst base classifier, i.e., the one with the largest bound. The search for the worst classifier $q^{*}$ may be formulated as the optimization problem:
\begin{equation} \label{eq:nabla_beta_h}
    \begin{aligned}
    q^{*} = &\argmax_{q \in \mathcal{Q}} \int q(y) \eta(y, \hat{x}) \rho(y \vert \hat{x}) d y, \\
    &\text{s.t.}\quad h (\hat x)=\int q(y) \rho(y \vert \hat{x}) d y,
    \end{aligned}
\end{equation}
where $\mathcal{Q} = \{q | q: X \to [0, 1]\}$ is the set of all binary classifiers. Under the specific choice of resolvable transform $\phi$ and perturbation distribution $\tau(\alpha)$, the problem from Eq. \eqref{eq:nabla_beta_h} admits the analytical solution.
In general, if the evaluation of $\rho(y|\hat x)$ and $\eta(y,\hat{x})$ are available, the solution of the problem from Eq. \eqref{eq:nabla_beta_h} could be obtained numerically. 

Namely, suppose that $M \in \mathbb{N}$ is the number of independent and identically distributed variables $(q_1,\eta_1), \dots, (q_M,\eta_M) \sim \rho(y|\hat x) \times \eta(y,\hat{x})$.  Then, the functional and constraint (respectively) from Eq. \eqref{eq:nabla_beta_h} can be approximated in the following form:
\begin{equation}\label{eq:fin_sample_for_density}
\begin{aligned}
    & \int q(y) \eta(y, \hat{x}) \rho(y|\hat{x}) dy \approx \frac{1}{M} \sum_{k=1}^M q_k \eta_k,
    \\
    & \int q(y) \rho(y|\hat{x}) dy \approx \frac{1}{M} \sum_{k=1}^M q_k.
\end{aligned}
\end{equation}

An approximate solution to Eq. \eqref{eq:nabla_beta_h} is then obtained by sorting $\eta_{i_1} \ge \dots \ge \eta_{i_M}$  and assigning 
\begin{equation}
\label{eq:numerical_ass}
    \begin{cases}
    q_{i_1}= q_{i_2} = \dots = q_{i_k} =  1, \\
    q_{i_{k+1}} = q_{i_{k+2}} = \dots = q_{M} = 0.
\end{cases}
\end{equation}
In the Eq. \eqref{eq:numerical_ass}, threshold index $k$ is chosen such that 
\begin{equation}\label{eq:empirical_inv_cdf}
\begin{cases}
    |h(\hat{x}) - S_k| \le |h(\hat{x}) - S_{k-1}|, \\
    |h(\hat{x}) - S_k| \le |h(\hat{x}) - S_{k+1}|,
\end{cases}
\end{equation}
where $S_k = \frac{1}{M} \sum_{j=1}^k q_{i_j}.$

For sufficiently large $M$, this scheme yields a tight approximation of $q^{*}$ from Eq. \eqref{eq:nabla_beta_h}, and, hence, for the bound $w(h, \beta)$ from Eq. \eqref{eq:bound_for_dd}. 

\subsubsection{Density Estimation} \label{section:332}

Since the exact evaluation of the density from Eq. \eqref{eq:density} is challenging, we emulate sampling from the conditional density $\rho(y|\hat{x})$ by estimating the gradient of the log-density $\eta(y, \hat{x})$ from Eq. \eqref{eq:nabla_h}.

Namely, we use the first-order approximation for the resolvable transform:
\begin{equation}
\begin{split}
    \phi(\hat x, \alpha) = & \phi\left(\hat x, \alpha_0\right) +J(\alpha_0)\left(\alpha-\alpha_0\right) + \\& + \mathcal{O}(\|\alpha-\alpha_0\|^2),       
\end{split}
\end{equation}
where  $J(\alpha) = \frac{\partial \phi}{\partial \alpha}$ 
to establish Laplace's posterior log-density estimation: 
\begin{equation}
\begin{aligned}
\label{eq:laplace_log_p}
\log \rho(y \vert \hat{x}) \approx & \log C  -\frac{\|\mu\|^2}{2 \sigma^2}+\left\langle M \alpha_0, \alpha_0\right\rangle - \\ &-\frac{1}{2} \log \det M,
\end{aligned}
\end{equation}
where $M = J^T J + \sigma^2 I$ and $\mu = y - \phi(\hat x, \alpha) + J\alpha$ and $C$ is a constant.
Finally, an approximation for the initial point $\alpha_0$ from the Eq. \eqref{eq:laplace_log_p} is given via one iteration of the Gauss-Newton method \cite{bjorck1996numerical}:
\begin{equation}
\alpha_0 \approx\left(J^{\top} J+\sigma^2 I\right)^{-1}(y-\phi(\hat{x}, \alpha)+J \alpha).
\end{equation}

While the above derivation admits an arbitrary parametric transform $\phi$, for the resolvable one, there exists a closed-form limit when $\sigma \to 0$. The last is summarized in the Lemma~\ref{lemma:logrho}, allowing us to compute log-density either analytically or through automatic differentiation tools.

\begin{lemma}{}
\label{lemma:logrho}
    Let $\gamma(\alpha, \beta)$ be the resolving function: $ \phi(\phi(x, \beta), \alpha) = \phi(x, \gamma(\alpha, \beta))$. Then, the formula for the logarithm of the conditional density from Eq. \eqref{eq:density}
   has the limit when $\sigma \rightarrow 0$ in the form
   \begin{equation}
        \log \rho(y \vert \hat{x}) = -\frac{1}{2} \log \det J^{\top} J + \log \tau(\alpha), \quad J = \frac{\partial \phi}{\partial \alpha}.
   \end{equation}
\end{lemma}

If only the log-density $\log \rho(y\vert \hat{x})$ is known, the expression for $\eta(y, \hat{x}) = \nabla_\beta \log \rho(y \vert \hat{x})$ is given by the following lemma:

\begin{lemma}[Gradient of log-density for resolvable transformations]
\label{lemma:resolvable}    
Suppose that the log-density $\log \rho(y| \hat x) = z(\alpha, \beta) = z(\alpha(\beta), \beta)$ is known.  Then 
\begin{equation*}
\eta(y \vert \hat{x}) = \nabla_{\beta} z = \frac{\partial z}{\partial \beta} - \frac{\partial z}{\partial \alpha} \left(\frac{\partial \gamma}{\partial \alpha}\right)^{\dagger} \frac{\partial \gamma}{\partial \beta},
\end{equation*}
where $\gamma$ is a resolving function of the transform: $\phi(\phi(x, \beta), \alpha) = \phi(x, \gamma(\alpha, \beta))$.
\end{lemma}
\begin{remark}
    Proofs of the lemmas are moved to the Appendix so as not to distract the reader.
\end{remark}
The overall procedure is presented in Algorithms~\ref{alg:main}, \ref{alg:comp_norm_bounds}.

\begin{algorithm}[htb]
\caption{Numerical Estimation of $\xi$ and $\hat g$ for the Resolvable Transform $\phi$}
\label{alg:main}
\begin{algorithmic}[1] 
\REQUIRE $\phi$ -- resolvable input transformation, \\
$N_{s}$ -- number of samples for bound estimation, \\
$\gamma$ -- resolving function of $\phi$, \\ 
$\beta_{0}$ -- identity parameters of $\phi$, \\
$\alpha$ -- smoothing parameter, \\
$B$ -- parametric grid of $d_{b}$ points to estimate bounds on, \\ 
$d$ --  number of parameters of the transform, \\
$x$ -- a random input point.
\ENSURE $\xi(h), \hat{g}(\beta)$ -- functions from Theorem \ref{theor:main}.

\STATE $\{p_{i}\}_{i=0}^{N_{s}}, \{g_{j}\}_{j=0}^{d_{b}} \gets \newline \gets \textsc{ComputeNormedBounds}(\gamma, \beta_{0}, B, N_{s}, \phi, d, x)$
\STATE $\{\xi_{i}\}_{i=0}^{N_{s}} \gets \frac{1}{N_s}\textsc{CumulativeSummation}\left(p_i^{-1}\right)$ 
\STATE $\xi(h) \gets \textsc{Interpolate}(\textsc{linspace}(0, 1, N_{s}), \xi_{i})$
\STATE $z \gets \textsc{Interpolate}\left(B, \{g_{j}\}_{j=0}^{d_{b}}\right)$
\FOR{$\beta_{j} \in B$}
    \STATE $\hat g_{j} = \int_0^1 z((1 - t)  \beta_0 + t  \beta_{j}) dt$
\ENDFOR
\STATE $\hat g(\beta) = \textsc{Interpolate}(B, \{\hat g_{j}\})$
\STATE \textbf{return} $\xi(h), \hat g(\beta)$

\end{algorithmic}
\vspace{1em}
\end{algorithm}

\begin{algorithm}[htb]
\caption{Compute Normed Bounds}
\label{alg:comp_norm_bounds}
\begin{algorithmic}[1]
\REQUIRE $\phi$ -- resolvable input transformation, \\
$N_{s}$ -- number of samples for bound estimation, \\
$\gamma$ -- resolving function of $\phi$, \\ 
$\beta_{0}$ -- identity parameters of $\phi$, \\
$\alpha$ -- smoothing parameter, \\
$B$ -- parametric grid of $d_{b}$ points to estimate bounds on, \\ 
$d$ --  number of parameters of the transform, \\
$x$ -- a random input point.
\ENSURE $\{p_i\}_{i=0}^{N_s}, \{g_j\}_{j=0}^{d_b}$ -- point-wise estimation of the gradient bound from Eq. \eqref{eq:bound_for_dd}.
    \FOR{$\beta_j \in B$}
        \STATE $c_j = \mathcal{N}(0, I_d)$
        \STATE $z \gets \textsc{LogRho}(x, \gamma, \beta_j, c_j, \phi, N_s)$ \COMMENT{Equation \eqref{eq:laplace_log_p}, Lemma~\ref{lemma:logrho}}
        \STATE $\eta \gets \textsc{GradLogRho}(x, \gamma, \beta_j, \alpha, \phi, N_s, z)$ \COMMENT{Equation \eqref{eq:laplace_log_p}, Lemma~\ref{lemma:resolvable}}
        \STATE $t \gets (\beta_j - \beta_0) / \| \beta_j-\beta_0 \|$
        \STATE $\eta \gets \eta t$
        \STATE $\eta \gets \eta - \textsc{Mean}(\eta)$
        \STATE $\eta \gets \textsc{Sort}(\eta, \text{reverse})$
        \STATE $\text{bound}_{j, :} \gets \textsc{CumulativeSummation}(\eta)/ N_s$
        \STATE $g_j \gets \max_i(\text{bound}_{j, i}) \|\beta_0 - \beta_j \|$
        \STATE $\text{bound}_{j, :}  \gets \text{bound}_{j, :} / \max_i(\text{bound}_{j, i})$
    \ENDFOR
    \STATE $p_i = \max_j{\text{bound}_{ji}}$
    \STATE \textbf{return} $\{p_i\}_{i=0}^{N_s}, \{g_j\}_{j=0}^{d_b}$
\end{algorithmic}
\end{algorithm}

\section{Experiments}

\begin{table*}[!h]
    \caption{\label{tab:main_table} Quantitative results on ImageNet dataset. We report smoothing distributions and certified robust accuracy for our approach and competitors' methods. The best results are highlighted in \textbf{bold}, \underline{underlined} denotes equivalent performance. Symbol $''-''$ in the table corresponds to the transformation in which a method does not certify the model against the given distribution parameters. We evaluate the CRA in the fixed parameter range $R_l ~\leq ~\beta ~\leq ~ R_r$ for each transformation type. In the parameter column, $c, ~ b, ~\gamma, ~ (T_x, T_y), ~r_b$ represent contrast, brightness, gamma-correction, translations, and Gaussian blur attacks' parameters, respectively. \text{CRA-TSS}, \text{CRA-MP}, and \text{CRA-GS} correspond to the certified accuracy of the methods from \cite{li2021tss}, \cite{muravev2021certified}, and \cite{hao2022gsmooth} respectively. The architecture of the base model is Resnet-50.}
    \label{table:main}
    \centering
    \begin{tabular}{lrccccccc}
        \toprule
       {Transform} & $\beta$ & $R_l$ & $R_r$ & {Distribution} & {CRA (ours)}& {CRA-TSS}  & {CRA-MP} & {CRA-GS} \\
        \midrule
        Brightness & $b$ & -0.4 & 0.4 & $\mathcal{N}(0, 0.3)$ & \textbf{0.69} & 0.68 & --& 0.67\\
        \midrule
        Contrast & $c$ & 0.6 & 1.4 & ${\rm LogNorm}(0, 0.3)$ & \underline{0.68} & -- & \underline{0.68} & 0.67\\
        \midrule
        Blur & $r_b$ & 1 & 4 & ${\rm Exp}(0.3)$ & \underline{0.59} & \underline{0.59} & --& 0.0\\
        \midrule
        Translation & $T_x, T_y$ & -56 & 56 & $\mathcal{N}(0, 50)$ & \textbf{0.49} & 0.28 & -- & 0.45\\
        \midrule
         Gamma & $\gamma$ & 1.0 & 2.0 & ${\rm Rayleigh}(0.1)$ & \textbf{0.66} & -- & 0.54 & --\\
         \midrule
         Gamma & $\gamma$ & 0.5 & 1.0 & ${\rm Rayleigh}(0.1)$ & \textbf{0.66} & -- & 0.61 & --\\
        \midrule
        Contrast & $c$ & 0.6 & 1.4 & \text{LogNorm}(0, 0.6) & \multirow{2}{*}{\underline{0.62}} & \multirow{2}{*}{0.59} & \multirow{2}{*}{--} & \multirow{2}{*}{\underline{0.62}}
        \\
        Brightness & $b$ & -0.4 & 0.4 & $\mathcal{N}(0, 0.6)$ \\
        \midrule
        Gamma & $\gamma$ & 0.8 & 1.4 & ${\rm Rayleigh}(0, 0.1)$ & \multirow{2}{*}{0.62} & \multirow{2}{*}{--} & \multirow{2}{*}{--} & \multirow{2}{*}{--}\\
        Contrast & $c$ & 0.6 & 2.0 & ${\rm LogNorm}(0, 0.1)$\\
        \midrule
        Brightness & $b$ & -0.2 & 0.2 & $\mathcal{N}(0, 0.4)$ & \multirow{2}{*}{\textbf{0.46}} & \multirow{2}{*}{0.02} & \multirow{2}{*}{--} & \multirow{2}{*}{--}\\
        Translation & $T_x, T_y$ & -56 & 56 & $\mathcal{N}(0, 30)$\\
        \midrule
        Contrast & $c$ & 0.8 & 1.2 & ${\rm LogNorm}(0, 0.4)$ & \multirow{2}{*}{0.09} & \multirow{2}{*}{--} & \multirow{2}{*}{--} & \multirow{2}{*}{--}\\
        Translation & $T_x, T_y$ & -25 & 25 & $\mathcal{N}(0, 30)$\\
        \midrule
        Contrast & $c$ & 0.8 & 1.2 & ${\rm LogNorm}(0, 0.4)$ & \multirow{3}{*}{0.06} & \multirow{3}{*}{--} & \multirow{3}{*}{--} & \multirow{3}{*}{--}\\
        Brightness & $b$ & -0.2 & 0.2 & $\mathcal{N}(0, 04)$\\
        Translation & $T_x, T_y$ & -15 & 15 & $\mathcal{N}(0, 15)$\\
        \midrule
         Translation & $T_x, T_y$ & -3 & 3 & $\mathcal{N}(0, 10)$ & \multirow{4}{*}{0.20} & \multirow{4}{*}{--} & \multirow{4}{*}{--} & \multirow{4}{*}{--}\\
         Blur & $r_b$ & 1 & 3 & ${\rm Rayleigh}(1)$\\
         Brightness & $b$ & -0.1 & 0.1 & $\mathcal{N}(0, 0.3)$\\
         Contrast & $c$ & 0.95 & 1.05 & ${\rm LogNorm}(0, 0.3)$\\
        \bottomrule
    \end{tabular}
    \vspace{0.5em}
\end{table*}

\label{s:experiments}
We conducted experiments with different ResNet architectures models on ImageneNet, CIFAR-10, and CIFAR-100 datasets\footnote{Our code is publicly available on \hyperlink{https://github.com/dkorzh10/general_lipschitz}{github.com/dkorzh10/general\_lipschitz}.}. The models were modified with an additional normalization layer as in \cite{cohen2019certified, li2021tss}. For a fixed type of semantic transformation $\phi$, we train base classifier $f$ with corresponding augmentation with the parameters sampled from the distribution mentioned in Table \ref{tab:main_table} to make the base classifier $f$ more empirically robust to this type of transformation. 
Depending on the transformation type, fine-tuning the pre-trained ImageNet models with augmentations takes from $3$ to $18$ hours on a $1$ Nvidia V-100 16GB GPU. The combination of the cross-entropy and consistency losses from 
\cite{jeong2020consistency} is chosen as an optimization objective, and the fine-tuning is conducted for $2$ epochs using SGD (with the learning rate of $10^{-3}$ and momentum of $0.95$). 

To evaluate our approach, we compute \emph{certified robust accuracy} (CRA) of the smoothed classifier. 
Certified robust accuracy is a fraction of correctly predicted images $x_i$ from the test set on which the certification condition is met.

\begin{figure*}[htb] 
\centering
\begin{subfigure}{0.33\textwidth} 
    \includegraphics[width=\textwidth]{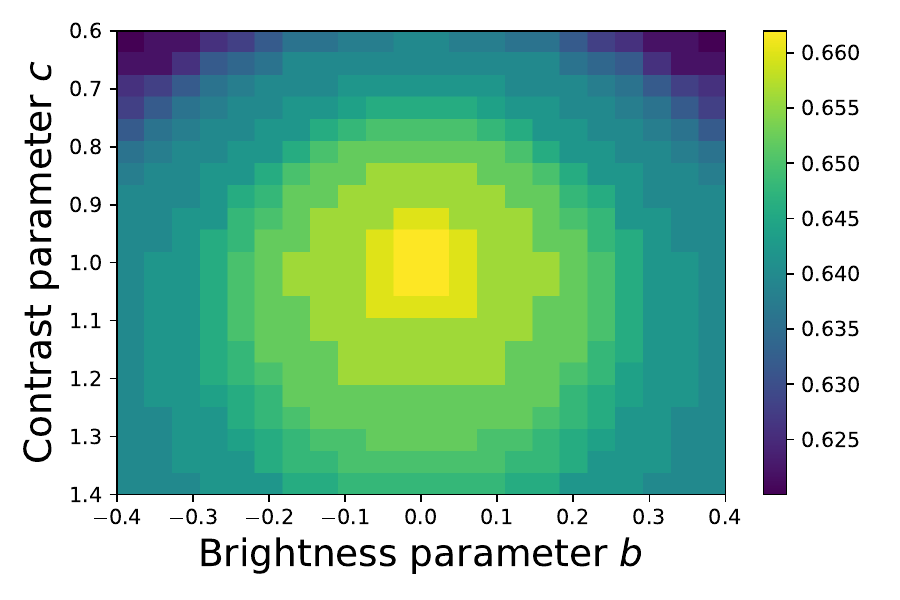}
    \caption{Contrast and Brightness}
    \vspace{1.5em}
\end{subfigure}
\begin{subfigure}{0.33\textwidth} 
    \includegraphics[width=\textwidth]{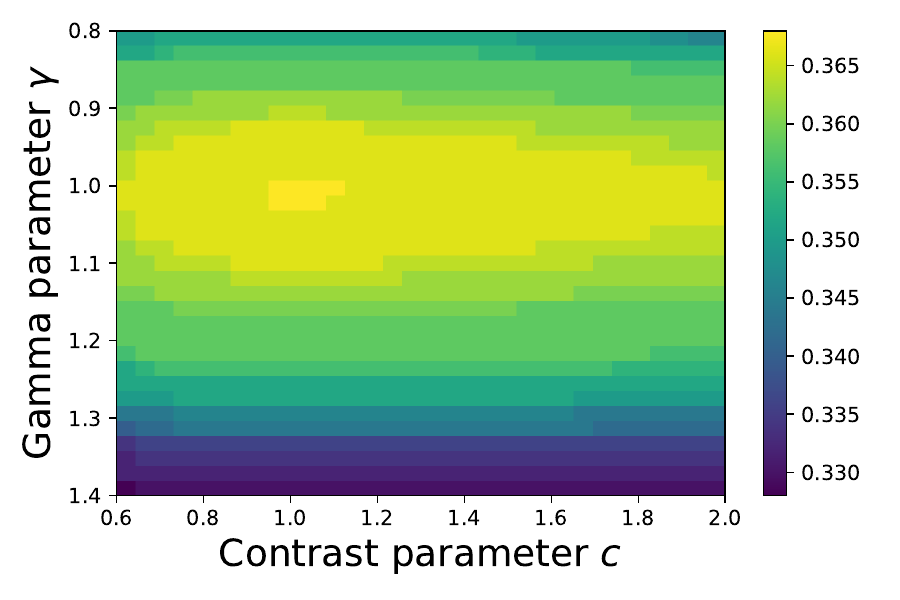}
    \caption{Gamma and Contrast}
    \vspace{1.5em}
\end{subfigure}
\begin{subfigure}{0.33\textwidth} 
    \includegraphics[width=\textwidth]{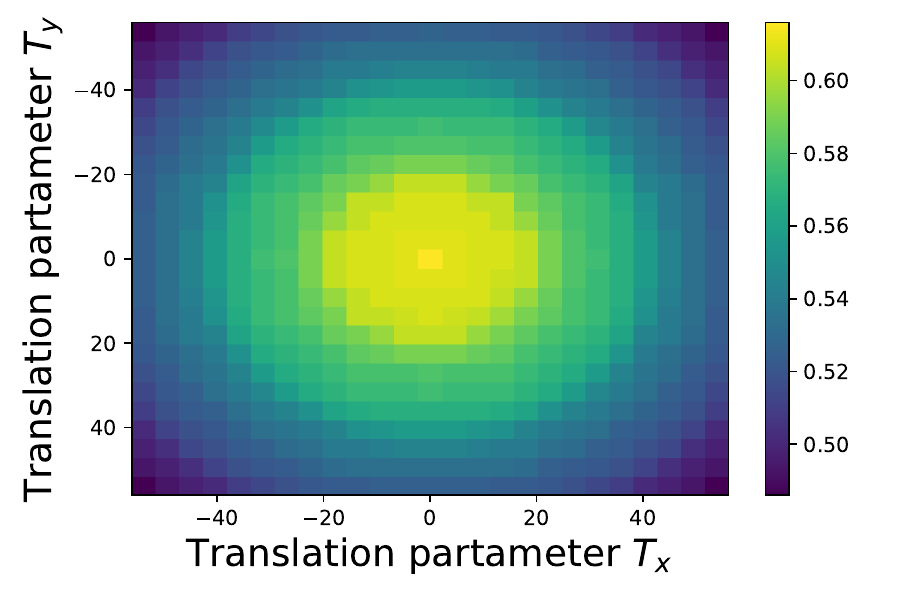}
    \caption{Translation}
    \vspace{1.5em}
\end{subfigure}
\caption{Visualization of certified robust accuracy for the subset of parameter space for different transformations, ImageNet dataset. By design of our approach, if the classifier is certified at the input point $x$ for the parameter value $\beta$, it is certified for all parameters $\beta^{*} \in [\beta_0, \beta].$ The values of CRA are presented in the corresponding color bars. Remark: the certified robust accuracy against the given transform in Table \ref{tab:main_table} is the infimum of CRAs on the corresponding plot.}
\label{fig:cra_transforms}
\vspace{0.5em}
\end{figure*}

CRA is evaluated on $500$ images sampled randomly from the test dataset. To estimate the prediction of the smoothed classifier $h$, we compute the lower bound of the Clopper-Pearson confidence interval \cite{clopper1934use} over the sample size $N_\text{max}=1000$ and confidence level $\alpha^* = 10^{-3}$ for each initial image $x.$ To estimate mappings $\xi, \hat g$, we sample parameters $\alpha$ from the Gaussian distribution and map them to the desired distribution (see Table \ref{tab:main_table}) using the numerical scheme. In our experiments, we sample parameters of additive transformations from Normal distribution, multiplicative transformations -- from LogNormal and Rayleigh distributions, and exponential transformations -- from Rayleigh distribution. In our approach, the certification procedure is sample-agnostic: it has to be done only once for a pair ``base network -- input transform''. Then, the certification in a new sample is done at the cost of one forward pass of the smoothed network. To estimate the computational complexity of the proposed method, we report the time required for the certification in Table \ref{tab:times}. Visualization of the results of the certification is presented in Figure \ref{fig:cra_transforms}.

\begin{table}[hpbt]
    \vspace{1em}
    \caption{Computation time in seconds for the certification, ImageNet dataset. We use $N_{max}=1000$ samples for smoothing. We report construction (constr) and certification (cert) time for our approach. 
    We measured the average time required to certify $500$ images for our method, TSS \cite{li2021tss} and MP \cite{muravev2021certified}.}
    \label{tab:times}
    \centering
    \begin{tabular}
    {lcccc}   
        \toprule
         \multirow{2}{*}{Transform} & Ours & Ours & \multirow{2}{*}{TSS} & \multirow{2}{*}{MP} \\
         & constr & cert &  &  \\
        \midrule
        Contrast Brightness &  170 &  1350 &  1675 & -- \\
        Brightness Translation &  200 &  2905 &  1500 & -- \\
        Translation & 33.2 & 1452 & 1505 & -- \\
        Gamma & 3.4 & 1450 & -- & 1470\\
        \bottomrule
    \end{tabular}
\end{table}

We evaluated our approach against \cite{li2021tss, muravev2021certified, hao2022gsmooth} and present the results in Tables \ref{tab:main_table} and \ref{tab:1d_cra}. 
Our method achieves state-of-the-art robustness certificates for the majority of transformations, such as Gamma-Contrast and Contrast-Translation.

{It is worth mentioning that for some compositions of transformations (namely, for ones involving both contrast adjustment in a wide range and image translations), the resulting classifier is ``over smoothed'' -- the estimation of probability $h_c(\hat{x})$ of ground truth class is often less than $0.5$. Hence, the necessary condition for our method's certification ($h_c(\hat{x}) > 0.5$) is often not satisfied, leading to underestimated CRA.} 

\begin{table}[htb]
    \caption{
        Certified robust accuracy (CRA) for some attacks on CIFAR-10 and CIFAR-100 datasets. The best results are highlighted in \textbf{bold}, \underline{underlined} denotes equivalent performance. For the contrast transform, our method vs. MP has $86.2$ vs. $86.2$ and $45.6$ vs. \textbf{46.0} for CIFAR-10 and CIFAR-100, respectively. {The architecture of the base model is Resnet-110.}}\label{tab:1d_cra}
    \vspace{1em}
    \centering
    \begin{tabular}
    {lcccccc}   
        \toprule
         \multirow{2}{*}{Transform} & \multicolumn{3}{c}{CIFAR-10} & \multicolumn{3}{c}{CIFAR-100} \\
         & Ours & TSS & GS  & Ours & TSS & GS\\
        \midrule
        Brightness & \textbf{86.8} & 86.6 & 85.6 & \textbf{45.6} & 43.8 &43.2 \\
        Contrast & \textbf{86.2} & -- & 85.6 & \textbf{45.6} & --& 43.2  \\
        Blur & 74.2 & \textbf{75.4} & 0.0  &  39.8 & \textbf{41.8}  & 0.0\\
        CB & \underline{85.5} & 83.4 & \underline{85.5} & \textbf{41.6} & 38.0& 41.4  \\
        \bottomrule
    \end{tabular}
    \vspace{0.5em}
\end{table}

\section{Limitations}

This section is devoted to discussing the limitations of the proposed approach.

\subsection{Non-resolvable transformations}
The major limitation of the proposed approach is that it is suitable to certify models only against resolvable perturbations. In the case of non-resolvable transformation, the conditional density from Eq. \eqref{eq:density} may not be a continuously differentiable function with respect to the transformation parameter in limit $\sigma \to 0$.

\subsection{Probabilistic certification}
Recall that our approach is based on the randomized smoothing technique; hence, the certified model can not be evaluated exactly. In our experimental setting, for the sample $x$ of class $c$, the true value of the smoothed classifier $h_c(x)$ is estimated as the lower bound of the Clopper-Pearson confidence interval \cite{clopper1934use} over $N_\text{max}$ samples for some confidence level $\alpha^*$. Namely, $\hat{h}(x) = B(\alpha^*/2, n, N_\text{max}-n+1),$ where $B$ is Beta distribution, $N_\text{max}$ is the sample size and $n$ is the number of perturbations for which $f(\phi(x, \alpha^j)) > \frac{1}{2}$. Thus, our approach produces certificates with probability $p\ge 1-\alpha^*,$ where $\alpha^*$ is the upper bound on the probability to return an overestimated lower bound for the value $h(x).$ For comparison, in our settings, we choose $\alpha^* = 10^{-3}$ and $N_\text{max}=1000.$

\subsection{Error Analysis}
\begin{figure}[b]
    \centering
    \includegraphics[width=0.40\textwidth]{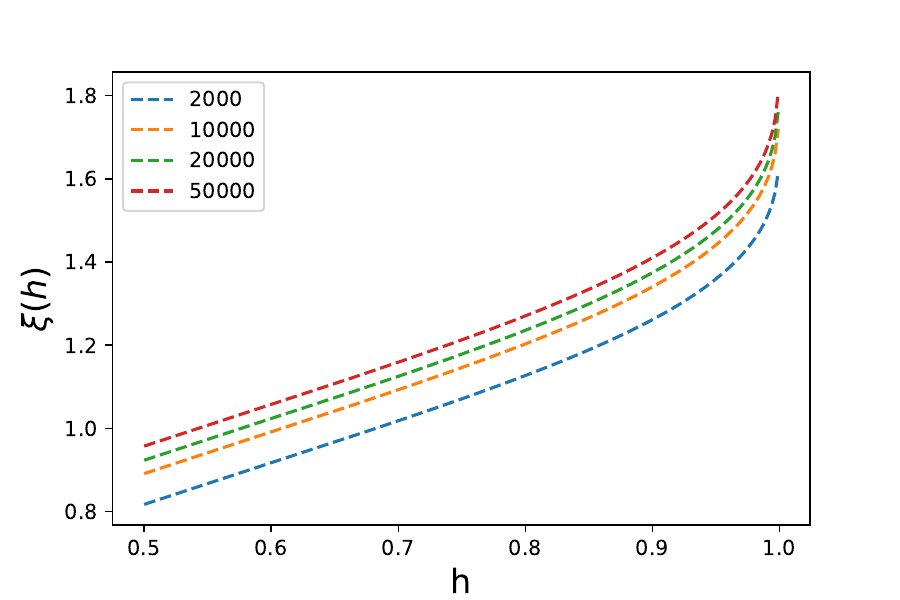}

    \caption{$\xi(h)$ v.s. $N_s$ for the Contrast-Brightness transform}
    \label{xi_vs_Ns_0}
    \vspace{1.5em}
\end{figure}

\begin{figure}[ht]
    \centering
    \includegraphics[width=0.40\textwidth]{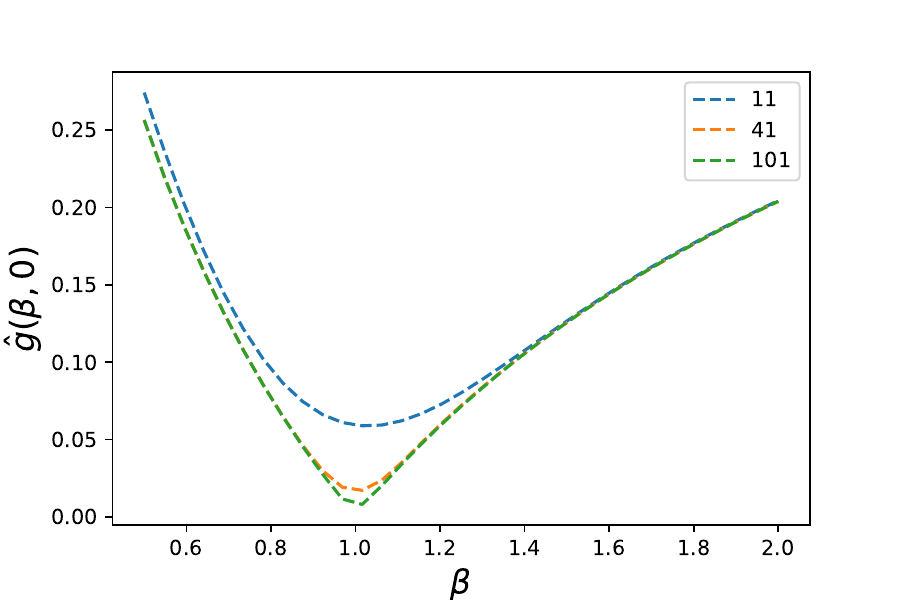}
    \caption{$\hat g(\beta, 0)$ v.s. $d_b$ for the Contrast-Brightness transform}
    \label{hatg_vs_db_0}
    \vspace{1.5em}
\end{figure}

 While certain transformations admit analytical solutions, numerical schemes inherently carry errors. However, attaining the desired precision is feasible by augmenting the sample size and refining the approximation with additional points \ref{xi_vs_Ns_0}, \ref{hatg_vs_db_0}. Considering these figures, it is visible that the impact of varying $N_s$ on the behavior of $\xi$ is relatively inconsequential for sufficiently large values. Similar observations can be made for $\hat g$. Assuming a fixed brightness parameter of $\beta_0$ and varying the contrast parameter, Figure \ref{hatg_vs_db_0} demonstrates that while differences exist, they remain insignificant. Analogous results may be anticipated by varying the brightness parameter instead of the contrast. 

 It is worth mentioning that the estimation of the error of empirical CDF and inverse empirical CDF \eqref{eq:numerical_ass}, \eqref{eq:numerical_ass}, 
 might be done, for example,  by applying Dvoretzky–Kiefer–Wolfowitz inequality.

The error estimation of numerical methods, particularly the more detailed estimation of the error of empirical cumulative distribution function and its inverse derived from numerical methods, is moved to the Appendix.
\section{Related Work}

\subsection{Adversarial attacks and empirical defenses}

The vulnerability of deep learning models to specifically crafted imperceptible additive transformations was discovered in \cite{szegedy2013intriguing, biggio2013evasion}. 
In \cite{goodfellow2014explaining}, 
the fast gradient sign method was proposed, and the empirical effectiveness of adversarial training against such transformations was discussed. 
Lately, several ways to exploit this vulnerability in the white-box \cite{moosavi2016deepfool} and black-box \cite{guo2019simple}
settings were proposed. Many applications of adversarial attacks were lately studied, especially in the real-world setting: face recognition \cite{komkov2021advhat,pautov2019adversarial}, detection \cite{kaziakhmedov2019real}, segmentation \cite{kaviani2022adversarial}, and other supervised settings were shown to be affected by this vulnerability. At the same time, empirical defenses \cite{madry2017towards, jia2022prior, ge2023towards} against such attacks were proposed. They are mainly narrowed to a specific attack setting or crafting the worst-case examples and including ones in the training set. 
However, such defenses did not provide enough guarantees to unseen attacks, and consequently, new adaptive attacks were created to overcome previously suggested defenses, causing an arms race.  
\par
\subsection{Approaches for provable certification}
Recent research in the field of certified robustness 
incorporates plenty of verification protocols \cite{li2020sok}. New approaches are often introduced during the competition on the verification of neural networks \cite{brix2023fourth}. There are two major approaches to certify classifiers against additive transformations: deterministic \cite{gowal2018effectiveness, wei2022certified} and probabilistic \cite{cohen2019certified, lecuyer2019certified}. The deterministic approach guarantees that the model is robust at some point $x$ if there is no point from the vicinity of $x$ such that it is classified differently from $x$. 

In contrast, probabilistic approaches are mainly based on randomized smoothing and utilize global \cite{gouk2021regularisation, tsuzuku2018lipschitz} or local \cite{wong2018provable} Lipschitz properties of the smoothed classifier. However, since a smoothed model can not be exactly evaluated, all the robustness guarantees hold with some probability depending on the finite sample estimation of the smoothed model \cite{cohen2019certified}.
Randomized smoothing is also applied in different domains, for example, as a defense against text adversarial attacks \cite{zhang2023text}, and automatic speech recognition defense \cite{olivier2021sequential}.
Solver-based deterministic approaches verify the model's robustness entirely but are not scalable due to computation complexity and usually are restricted to simple architectures \cite{katz2019marabou}. In contrast, linear relaxation approaches do not provide the tightest possible robustness certificates but are model-agnostic and applicable to large datasets \cite{wong2018provable}. The other deterministic methods are usually based on particular properties of neural networks, such as Lipschitz continuity \cite{levine2021improved,tsuzuku2018lipschitz} or curvature of the decision boundary \cite{singla2020second}. On the other hand, probabilistic approaches usually utilize an assumption about the smoothness of 
the model and provide presented state-of-the-art certification results against additive transformations \cite{cohen2019certified,lecuyer2019certified}.



Semantic transformations are another important class of input perturbations, which fool deep learning models easily \cite{ghiasi2020breaking, joshi2019semantic}. Certified robustness under this threat model is still an open issue \cite{li2021tss,alfarra2022deformrs}. Recently, a few approaches to tackle semantic transformations were proposed that are based on enumeration \cite{pei2017towards},  interval bound propagation \cite{balunovic2019certifying,mohapatra2020towards}, and randomized smoothing \cite{li2021tss,hao2022gsmooth}. It is known that different relaxation approaches provide worse results than the ones based on smoothing \cite{zhang2022rethinking}. On the other hand, when deterministic guarantees are infeasible, probabilistic approaches \cite{pautov2022cc,wicker2020probabilistic} to estimate the probability of the model failing when an attack is parameterized provide some insights about the model's robustness. It is worth mentioning that empirical robustness \cite{tsai2022towards} may be improved by incorporating adversarial training, which might be time-consuming on large-scale datasets \cite{tsai2022towards}. 



A promising way to tackle the certified robustness against semantic perturbations is based on transformation-specific randomized smoothing \cite{li2021tss,hao2022gsmooth}. The idea of transformation-specific smoothing is to consider the Lipschitz continuity of a smoothed model with respect to the transformation parameters. According to \cite{li2021tss}, this type of smoothing may be applied to two categories of transformations: \textit{resolvable} and \textit{differentially resolvable}, where the last implies taking into account interpolation errors. However, previous attempts to apply transformation-specific smoothing to certify classifiers against semantic transformations might be infeasible for more complicated
transformations \cite{li2021tss} or require a surrogate network to represent the transformation and not scalable to large datasets \cite{hao2022gsmooth}. 
As a separate application of transformation-dependent smoothing, \cite{muravev2021certified} specifically studied  Gamma correction and Contrast change as multiplicative transformations. This work provides asymmetrical guarantees and estimates the certification quality, considering realistic image compression into 8-bit RGB.
\section{Conclusion and future work}

In this paper, we propose General Lipschitz, a novel framework to certify neural networks against resolvable transformations and their compositions. Based on transformation-dependent randomized smoothing, our approach yields robustness certificates for complex parameterized subsets of parameter space. One of the advantages of the framework is the numerical procedure that produces certificates for a parameter subset by verifying certification condition in a single point of parameter space. Our experimental study shows that the proposed method achieves certified robust accuracy comparable to the state-of-the-art techniques and outperforms them in some experimental settings. Our approach allows us to certify models against resolvable transformations only, so one possible direction for future work is to extend it to the case of differentially resolvable transformations. Another direction is to apply the approach to object detection or segmentation tasks.



\newpage
\bibliography{mybibfile}

\newpage
\appendix
\onecolumn
\section{Appendix}

In this section, we present the proofs of theoretical results. 

\begin{theorem}{Certification condition (restated).}

Let $\beta(t): [0,1] \to \Theta$ be a smooth curve such that $\beta(0) = \beta_0$ and $\beta(1) = \beta$. Then there exist mappings $\xi: \left[0, 1\right] \to \mathbb{R}$ and  $\hat g(\beta): \Theta \to \mathbb{R}$ such that if 
$\hat g(\beta) < -\xi(1 -~h_c(x)) + \xi(1/2) $, then $h$ is robust at $x$ for all $\beta \in \beta(t), $ where $t \in [0, 1]$.

\end{theorem}

\begin{proof} 
Recall that a smoothed classifier has the form of expectation over perturbation density conditioned on the observed sample $x$, namely,

\begin{align}
    h(\phi(x, \beta)) \equiv h(\hat x) &\equiv h(x, \beta) \\ & \equiv h(\beta) = \int_{\mathbb{R}^n} f(y) \rho(y|\hat{x})dy,
\end{align}
where $\rho(y|\hat{x})$ is from Eq. \eqref{eq:density}. 

Let now $x$ be a fixed input object corresponding to class $c$. Since now, we have reassigned $h(\beta) = h_c(\beta)$ to manipulate only with $c-$~th component of the smoothed classifier. 
Let also $x$ be correctly classified with high confidence, namely, let $h(\beta) > \frac{1}{2}.$

To construct the certification criteria, we first observe that a directional derivative of $h(\beta)$ with respect to $\beta$ is bounded by the product of two functions, namely $p: [0,1] \to \mathbb{R}$ and $g: \Theta \to \mathbb{R}_{\ge 0}$ such that 

\begin{equation}
   \langle \nabla_\beta\  h(\beta), u \rangle = \int_{\mathbb{R}^n} f(y) \langle \nabla_\beta \rho(y|\hat{x}),u \rangle dy \le \tilde g(h, \beta) \le p(h(\beta)) g(\beta) 
\end{equation}
for all $u: \|u\|_2=1$. Note that such $p(\cdot)$ and $g(\cdot)$ exist since $h(\cdot)$ is assumed to be smooth (e.g., $p \equiv 1, g(\beta) \equiv \sup_{u} \sup_{\beta} \langle \nabla_{\beta} h(\beta), u\rangle$).

For the expression of the gradient, we have 
\begin{equation}\label{eq:nabla_beta_proof}
    \langle\nabla_\beta h(\beta), u \rangle = \int_{\mathbb{R}^n} f(y) \eta(y, \hat{x}) \rho(y|\hat{x}) dy,
\end{equation}
where $\eta(y, \hat{x}) = \langle \nabla_\beta \log \rho(y|\hat{x}), u \rangle$ and $u$ is fixed.

To estimate the supremum in Eq. \eqref{eq:nabla_beta_proof} given $\beta$, we solve an optimization problem in the form 

\begin{equation} \label{eq:nabla_beta_h_2}
\begin{aligned}
    & \argmax_{q \in \mathcal{Q}} \int_{\mathbb{R}^n} q(y) \eta(y, \hat{x}) \rho(y | \hat{x}) d y, 
    \\
    & \text { s.t. } \quad h=\int_{\mathbb{R}^n} q(y) \rho(y | \hat{x}) d y,
\end{aligned}
\end{equation}
where $\mathcal{Q} = \{q: X \to [0,1]\}$ is the class of base classifiers that satisfy the constraint in Eq. \eqref{eq:nabla_beta_h_2}. In other words, solving the problem in Eq. \eqref{eq:nabla_beta_h_2}, we find the worst-case base classifier $q$, i.e., the one with the maximum gradient value given $\beta.$

Note that the problem in Eq. \eqref{eq:nabla_beta_h_2} is known \cite{yang2020randomized} to have the optimal solution in the form 
\begin{equation}\label{eq:opt_sol}
    q^{*}(y) = \begin{cases} 1, & \text{if } \eta(y, \hat{x}) > F^{-1}_{\eta(y,\hat{x})} (1-h)\\
                     0, &  \text{otherwise},
       \end{cases}
\end{equation}
where $F^{-1}$ is the inverse CDF of random variable $\eta(y, \hat{x}).$
Hence, 
\begin{equation} \label{eq:hat_g}
\begin{aligned}
    & \int_{\mathbb{R}^n} q^{*}(y) \eta(y, \hat{x}) \rho(y|\hat{x})dy = 
    \\
    & \int_{z \in \mathbb{R}^n: \eta(z, \hat{x})> F^{-1}_{\eta(z,\hat{x})} (1-h)} \eta(z, \hat{x}) \rho(z|\hat{x})dz :=  \tilde{g}(h, \beta).   
\end{aligned}
\end{equation}
Thus, we have 
\begin{equation}\label{eq:what_to_int}
    \langle \nabla_\beta h(\beta), u \rangle \le  \tilde{g}(h, \beta),
\end{equation}
where the last quantity is bounded from above:
\begin{equation}
\begin{aligned}
    \tilde g(h, \beta) &= \frac{\tilde g(h, \beta)}{\sup_{h} \tilde g(h, \beta)} \sup_h \tilde g(h, \beta)\\
    &\leq \sup_{\beta} \left( \frac{\tilde g(h, \beta)}{\sup_{h} \tilde g(h, \beta)} \right) \sup_h \tilde g(h, \beta) = p(h) {g}(\beta).
\end{aligned}
\end{equation}
Now, integrating inequality from Eq. \eqref{eq:what_to_int} along a smooth curve $\beta(t): \beta(0) = \beta_0, \beta(1)=\beta$, we get 

\begin{equation}\label{eq:how_xi}
    \int_{\beta(t)} \langle \nabla_\beta h(\beta), u\rangle dt \le \int_{\beta(t)} p(h) g(\beta) dt.
\end{equation}
Introducing an auxiliary function $\xi(h) = \int \frac{1}{p(h)}dh,$ we get 
\begin{equation}\label{eq:bound_fin}
\xi(h(\beta)) - \xi(h(\beta_0)) = \int_{\beta(t)} \langle \nabla_{\beta} \xi(h(\beta)), u\rangle dt \le \int_{\beta(t)}g(\beta)dt = \hat{g}(\beta).
\end{equation}
Note that $\xi$ is a monotonically increasing function w.r.t. $h(\beta)$ according to the definition (i.e. $\rho(h) \ge 0$); and $\hat{g}(\beta)$ is non-decreasing  along $\beta(t)$ since $g(\beta)$ is non-negative. Notice now that inequality from Eq. \eqref{eq:bound_fin} holds for all components of the smoothed classifier since the right-hand side does not depend on $h$.


To consider the worst-case scenario (namely, the largest runner-up component of the classifier), suppose that the smoothed classifier is binary (without loss of generality, it corresponds to the one-vs-all classification scenario for the class $c$); we observe its component $h_{\ne c}(\beta)$ that corresponds to the probability of assigning a sample  \emph{not} to class $c$. 

For adversarial $\beta$, such that $h_{\ne c}(\beta) > 1/2$, we have

\begin{equation}
    \xi(h_{\ne c}(\beta)) > \xi(1/2), \quad \xi(h_{\ne c}(\beta_0)) = \xi(1 - h_{ c}(\beta_0))
\end{equation}
implying 
\begin{equation}\label{eq:crit_for_adv}
    \xi(1/2) - \xi(1 - h_c(\beta_0)) < \xi(h_{\ne c}(\beta)) - \xi (h_{\ne c}(\beta_0)) \le \hat{g}(\beta).
\end{equation}
Consequently, if the inequality in Eq. \eqref{eq:crit_for_adv} does not hold, it implies that the value $\beta$ in it is not adversarial. 
In other words, 
\begin{equation}
    \xi(1/2) - \xi(1 -h_c(\beta_0)) > \hat{g}(\beta)
\end{equation}
is the sufficient condition to certify $h$ for the value $\beta$ of the transformation parameter. Finally, recalling that $\hat{g}(\beta)$ is increasing along $\beta(t)$ finalizes the proof.

\end{proof}

\begin{lemma}
{Gradient of log-density for resolvable transformations} 
    
    If we have  log-density  \eqref{eq:density} $\log \rho(y| \hat x) = z(\alpha, \beta) = z(\alpha(\beta), \beta)$, 
    then $ \nabla_{\beta} z = \frac{\partial z}{\partial \beta} - \frac{\partial z}{\partial \alpha} \left(\frac{\partial \gamma}{\partial \alpha}\right)^{\dagger} \frac{\partial \gamma}{\partial \beta}, $ where $\gamma$ is a resolving function of the transform: $\phi(\phi(x, \beta), \alpha) = \phi(x, \gamma(\alpha, \beta))$.
\end{lemma}
\begin{proof}

Recall that $\log \rho(y| \hat x) = z(\alpha(\beta), \beta)$ and $y = \phi(\phi(x,\beta), \alpha) = \phi(x, \gamma(\alpha(\beta), \beta)).$
For the expression of the differential of $z$, we have
\begin{equation}
    dz = \frac{\partial z}{\partial \beta} d \beta + \frac{\partial z}{\partial \alpha} d \alpha.
\end{equation}
Consequently, the gradient of $z$ w.r.t $\beta$ has the form
\begin{equation}
    \frac{dz}{d \beta} = \frac{\partial z}{\partial \beta}+ \frac{\partial z}{\partial \alpha} \frac{d \alpha}{d \beta}.
\end{equation}
Analogously, for $\gamma = \gamma(\alpha(\beta), \beta),$   
\begin{equation}
    \frac{d \gamma}{d \beta} = \frac{\partial \gamma}{\partial \beta}+ \frac{\partial \gamma}{\partial \alpha} \frac{d \alpha}{d \beta}.
\end{equation}
Since we want to evaluate $\frac{dz}{d\beta}$, fixing $y$ implies $\gamma = \gamma(\alpha(\beta), \beta)) = \gamma_0$ to be constant. Thus,
\begin{equation}
    0 = d \gamma  = \frac{\partial \gamma}{\partial \beta} d \beta + \frac{\partial \gamma}{\partial \alpha} d \alpha.
\end{equation}
Hence, a gradient w.r.t. $\beta$%
\begin{equation}
    \frac{\partial \gamma}{\partial \beta} + \frac{\partial \gamma}{\partial \alpha} \frac{d \alpha}{d \beta} = 0
\end{equation}
implies 
\begin{equation}
    \frac{d \alpha}{d \beta} = - \left(\frac{\partial \gamma}{\partial \alpha}\right)^{\dagger} \frac{\partial \gamma}{\partial \beta}.
\end{equation}
As a result,
\begin{equation}
    \nabla_{\beta} \log \rho(y | x) = \frac{dz}{d \beta} = \frac{\partial z}{\partial \beta} - \frac{\partial z}{\partial \alpha} \left(\frac{\partial \gamma}{\partial \alpha}\right)^{\dagger} \frac{\partial \gamma}{\partial \beta}.
\end{equation}
\end{proof}


\subsection{Logarithm of conditional density for a resolvable transformation.}
\begin{lemma}{}
\label{lemma:logrho_bla}
    Let $\gamma(\alpha, \beta)$ be the resolving function:
    
   \begin{equation}
       \phi(\phi(x, \beta), \alpha) = \phi(x, \gamma(\alpha, \beta)).
   \end{equation}
   Then, the formula for the logarithm of the conditional density from Eq. \eqref{eq:density}
   has the limit when $\sigma \rightarrow 0$ in the form
   \begin{equation}
   \begin{aligned}
        & \log \rho(y \vert \hat{x}) = z(\alpha, \beta) = -\frac{1}{2} \log \det J^{\top} J + \log \tau(\alpha),
        \\
        & J = \frac{\partial \phi}{\partial \alpha}.
   \end{aligned}
   \end{equation}

\end{lemma}
\begin{proof}
    When $\sigma \to 0$, the limit of normal density from which comes noise $\varepsilon$ is the Dirac delta function. 
    \begin{align*}
        \rho(y \vert \hat{x}) 
        &= \int\limits_{\Theta}\delta(y - \phi(\hat{x}, \alpha))\tau(\alpha)d\alpha\\
        &= \tau(\phi^{-1}(y, \hat x)) \left(\det\left(J^{\top}J\right) \right)^{-\frac{1}{2}} 
    \end{align*}
\end{proof}



\subsection{Certified Radii Estimation Example}

Here, we present an analytical derivation for an additive perturbation.
$\phi(x, \alpha) = x + \alpha, ~~$
$\tau(\alpha) = \frac{1}{\sqrt{2 \pi \kappa^2}} e^{- \frac{\alpha^2}{2 \kappa^2}}$
\begin{equation}
\begin{split}
    & \frac{1}{\sqrt{2 \pi \sigma^2}}\int e^{-\frac{\|y-\phi(\hat{x}, \alpha)\|^2}{2 \sigma^2}} \tau(\alpha) d \alpha = 
    \\
    & \frac{1}{\sqrt{2 \pi \sigma^2}} \int e^{-\frac{\|y -(x + \alpha + \beta)\|^2}{2 \sigma^2}} \tau(\alpha) d \alpha = 
    \\
    &\frac{1}{\sqrt{2\pi }\sqrt{\sigma^2 + \tau^2}} e^{-\frac{(y - x -\beta)^2}{2(\sigma^2 + \kappa^2)}}
\end{split}
\end{equation}
The expression for the log density takes the form

\begin{equation}
    \log \rho(y \mid \hat{x})= -\frac{1}{2} \log(2 \pi) - \frac{1}{2} \log(\kappa^2 + \sigma^2)  - \frac{(y-x-\beta)^2}{2(\sigma^2 + \kappa^2)}
\end{equation}
    The same result for $\log \rho(y \mid \hat{x})$ might be obtained by applying Lemma \ref{lemma:logrho_bla}.
\begin{equation}
    \eta(y, \hat x) = \frac{\partial}{\partial \beta} \log \rho(y \mid \hat x) = \frac{y -x -\beta}{\sigma^2 + \kappa^2},
\end{equation}

\begin{equation}
    \frac{\partial h}{\partial \beta}=\int f(y) \frac{\partial}{\partial \beta} \rho(y \mid \hat{x}) d y=\int f(y) \eta(y, \hat{x}) \rho(y \mid \hat{x}) d y
\end{equation}
Now, we want to solve the problem from Eq. \eqref{eq:nabla_beta_h}. The optimal $f$'s values might be either $0$ or $1$:
\begin{equation}
    f = H(y-q) \quad \text{or} \quad  f = 1 - H(y-q),
\end{equation}
or linear combination of them, where $H$ is the Heaviside function and $q$ is an unknown threshold parameter.
Because $h(\beta )\in [0,1]$ in constraint, and following the numerical algorithm idea, we can realize that we have to "sum up" enough probability over probability density function $\rho(y| \hat x)$ or following the proof of \ref{theor:main}, find an optimal solution via inverse CDF of $\eta$ \ref{eq:opt_sol}. Thus,
\begin{equation}
    \int_q^{\infty} \rho(y| \hat x) dy =  \int_q^{\infty} \frac{1}{\sqrt{2 \pi}}\frac{1}{\sqrt{\sigma^2 + \kappa^2}} e^{\frac{-(y - x - \beta)^2}{2(\sigma^2 + \kappa^2)}}= h
\end{equation}
\begin{equation}
    \frac{1}{2}\left[ 1 + \operatorname{erf} \left( \frac{\beta - q + x}{\sqrt{2}\sqrt{\sigma^2 +\kappa^2}} \right)\right] = h,
\end{equation}
where $\operatorname{erf}$ is an error function. Applying the fact that $\operatorname{erf}^{-1}(x) = -\operatorname{erf}^{-1}(-x)$, obtain
\begin{equation}
    q = x + \beta + \sqrt{2}\sqrt{\sigma^2 + \kappa^2}\operatorname{erf}^{-1} \left( 1- 2h\right)
\end{equation}
If we integrate from $-\infty$ to $q$, we obtain $q = x + \beta - \sqrt{2}\sqrt{\sigma^2 + \kappa^2}\operatorname{erf}^{-1} \left( 1- 2h\right)$. Now, let's estimate the bound:
\begin{equation}
\begin{aligned}
    & \int_q^{\infty} \frac{y -x -\beta}{\sigma^2 + \kappa^2} \frac{1}{\sqrt{2 \pi} \sqrt{\sigma^2 + \kappa^2}} e^{-\frac{(y - x- \beta)^2}{2(\sigma^2 + \kappa^2)}} d \alpha =
    \\
    & = \frac{1}{(\sigma^2 + \kappa^2)^{\frac{3}{2}}\sqrt{2 \pi}} \int_{\tilde q}^{\infty} \alpha e^{- \frac{\alpha^2}{2 (\sigma^2 + \kappa^2)}} d\alpha  
\end{aligned}
\end{equation}
where $\tilde q = q - x -\beta$ due to the variables change. Integrating, receive
\begin{equation}
    \frac{\partial h}{\partial \beta} \leq g(\beta) p(h) = \frac{1}{\sqrt{\sigma^2 + \kappa^2}\sqrt{2 \pi}} e^{(\operatorname{erf}^{-1}(1-2h))^2}
\end{equation}
and (we can use all of this as a bound or split into $g(\beta)$ and $p(h)$, where $g(\beta)=1$. Finally,
\begin{equation}
\begin{aligned}
    & \xi(h) = \int \frac{1}{p(h)} dh = \sqrt{\sigma^2 + \kappa^2} \sqrt{2}\operatorname{erf}^{-1}(2h - 1)
    =
    \\
    & = \sqrt{\sigma^2 + \kappa^2} \Phi^{-1}(h),
\end{aligned}
\end{equation}
where $\Phi^{-1}$ is an inverse cumulative density function of Normal distribution. 


\subsection{Additional Experiments}
This section revisits the experimental aspects of our proposed numerical certification algorithm and delves into several crucial hyperparameters that influence its performance.

\begin{itemize}
    \item $N_s$ -- number of samples from $\eqref{eq:density}$, used for estimation of $\xi$ from \eqref{alg:comp_norm_bounds};
    \item $d_b$ -- number grid points employed for each dimension of the attack to estimate the bounds  $p(h), ~g(\beta)$, $\xi(h)$, and $\hat g(\beta)$ consequently.
    \item The number of grid points utilized in each dimension to estimate the Certified Radii of Applicability (CRA).
    \item Number of samples deployed for estimating the smoothed classifier using the Monte-Carlo method and Clopper-Pearson test.
\end{itemize}

\begin{figure}[h!]
    \centering
    \includegraphics[width=0.42\textwidth]{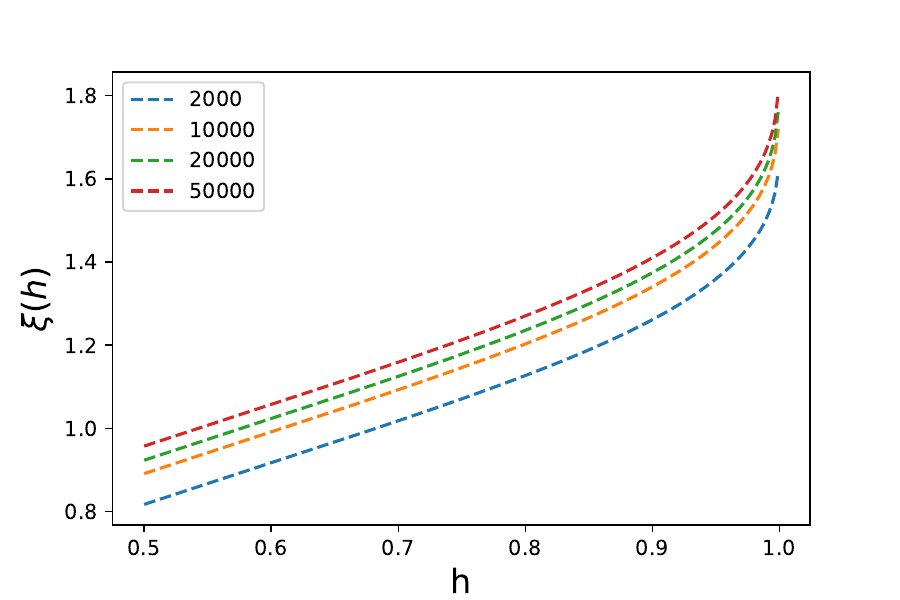}

    \caption{$\xi(h)$ vs $N_s$ for the Contrast-Brightness transform}
    \label{xi_vs_Ns}
    \vspace{2em}
\end{figure}

\begin{figure}[h!]
    \centering
    \includegraphics[width=0.42\textwidth]{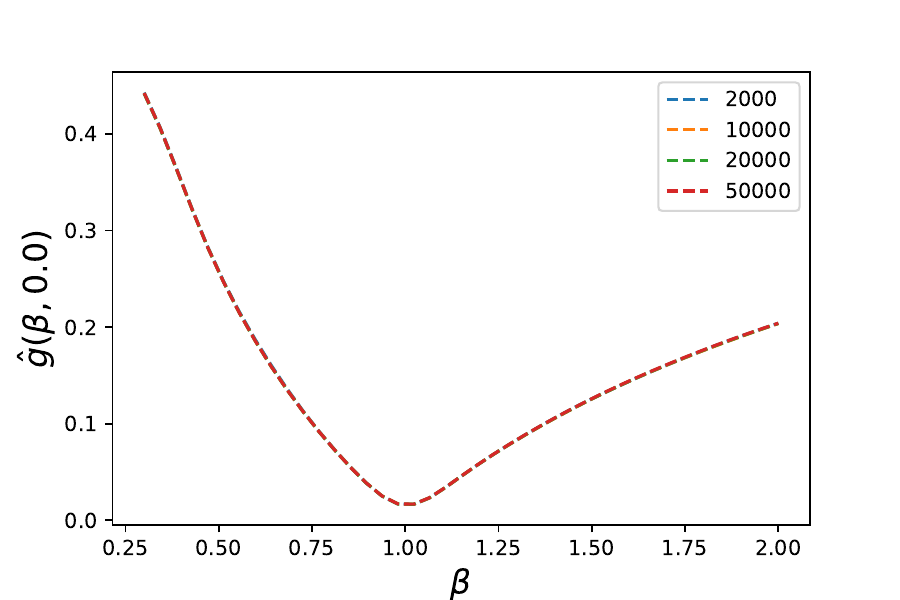}
    \caption{$\hat g(\beta, 0)$ vs $N_s$ for the Contrast-Brightness transform}
    \label{hatg_vs_Ns}
    \vspace{2em}
\end{figure}

Let us consider the Contrast-Brightness transform as an illustrative example. Examining Figure \ref{xi_vs_Ns}, it becomes apparent that the impact of varying $N_s$ on the behavior of $\xi$ is relatively inconsequential for sufficiently large values. 
Similar observations can be made for $\hat g$. Assuming a fixed brightness parameter of $\beta_0$ and varying the contrast parameter, Figure \ref{hatg_vs_Ns} demonstrates that while differences exist, they remain insignificant. Analogous results may be anticipated by varying the brightness parameter instead of the contrast.
\begin{figure}[h!]
    \centering
    \includegraphics[width=0.42\textwidth]{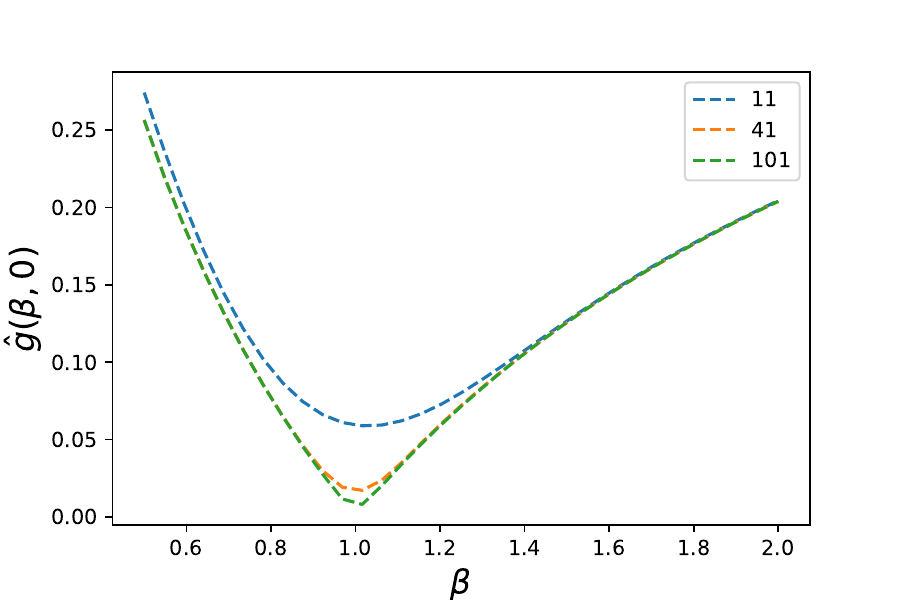}
    \caption{$\hat g(\beta, 0)$ vs $d_b$ for the Contrast-Brightness transform}
    \label{hatg_vs_db}
    \vspace{2em}
\end{figure}

On the other hand, if we vary $d_b,$ we can notice different effects for $\hat g$ (Figure \ref{hatg_vs_db}).
Notably, larger values of $\hat g$ are obtained with fewer points used for approximation (at least in this example). This fact supports the effectiveness of our numerical certification approach as empirical $\hat g$ gives an upper bound on the actual function, consequently enabling the valid application of the certification criteria \ref{theor:main}. The situation remains the same if we vary the brightness parameter.

\begin{figure}[h!]
    \centering
    \includegraphics[width=0.42\textwidth]{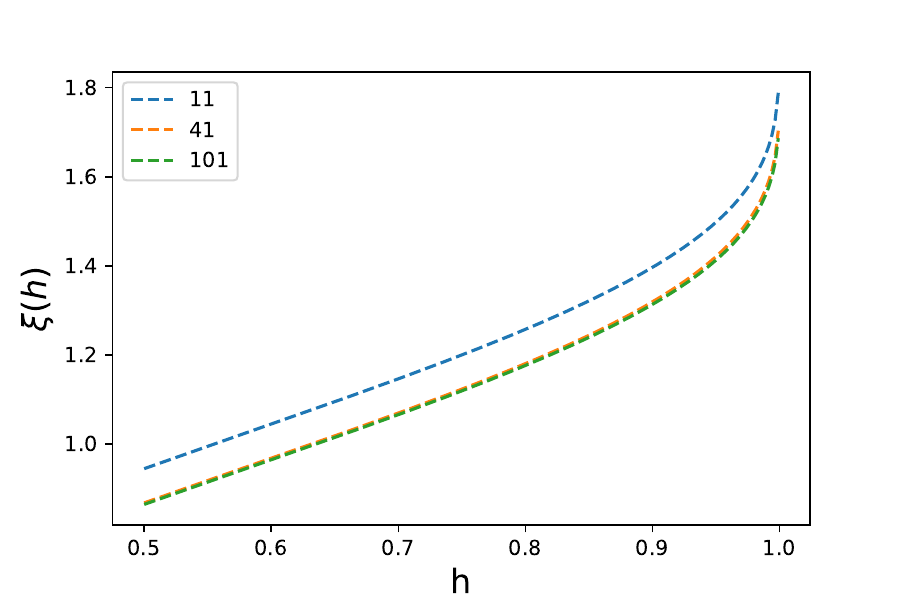}
    \caption{$\xi(h)$ vs $d_b$ for the Contrast-Brightness transform}
    \label{xi_vs_db}
    \vspace{2em}
\end{figure}

The situation is different for $\xi(h)$, as it produces higher results for worse approximation \ref{xi_vs_db}. Nonetheless, this discrepancy becomes less significant when evaluating certified radii, particularly when contrasting the difference $-\xi(1-h) + \xi(h)$, stabilizing with increasing $d_b$. 

The impact of attack grid size on CRA estimation appears negligible, as our investigation suggests. Specifically, there is minimal variation observed across $11$, $41$, and $101$ points per axis. This can be explained by acknowledging that the worst results should be observed on the borders of attacks due to the monotonic effects of our functions. 

The usage of quite a small number of samples ($N_{\text{max}}=1000$) is motivated by the negligible difference in certificates for~$N_{\text{max}}=1000$,~$10000$, and~$100000$ for almost all values of top-class probability of the smoothed classifier.
Our ImageNet models have this robustness-accuracy trade-off. Thus, we utilize~$n = 1000$.
However, we agree that the tightness of the certification significantly depends on the~$n$ when~$0.98\leq h\leq 1$ for the confident, smoothed classifiers. 



\section{Error Analysis of Numerical Methods}

In this section, we conduct an error analysis of our numerical certification procedure to demonstrate the feasibility of tight and reliable approximation.

Equations \eqref{eq:xi_and_hat_g},\eqref{eq:fin_sample_for_density}, \eqref{eq:empirical_inv_cdf}, \eqref{eq:laplace_log_p} should be assessed with corresponding sections and algorithms. First of all, we should emphasize that estimation of $\log \rho(y \vert \hat x)$ and $\eta(t, \hat x)$ is precise due to the lemmas \ref{lemma:logrho_bla} and \ref{lemma:resolvable}. Although we provide the result for non-zero sigma in section \ref{section:332}, for our experiments with resolvable transforms, $\sigma=0$  is applicable. The Gaussian smoothing with a non-zero standard deviation $\sigma$  makes the conditional density $\rho$ from Eq. \eqref{eq:density} an everywhere smooth function; this property may, in principle, be used for the certification of neural networks against non-resolvable transformations.

It is worth mentioning that the estimation of the error of inverse empirical CDF \eqref{eq:numerical_ass}, \eqref{eq:numerical_ass}, is vital for the precise computation of functions  $p(h)$ and $g(\beta)$ (and, hence, the values  $\xi(h)$ and $\hat g(\beta)$). 
The analytical CDF and its inverse are introduced in the proof of the theorem \ref{theor:main}: $F^{-1}_{\eta(y,\hat{x})}$. We need to estimate the convergence of empirical CDF $F^{-1}_{M, \eta(y,\hat{x})}$. One way to do it is to apply Dvoretzky–Kiefer–Wolfowitz inequality
\begin{equation}
\mathbb{P}\left(\sup _{x \in \mathbb{R}}\left|F_M(x)-F(x)\right|>\varepsilon\right) \leq 2 e^{-2 M \varepsilon^2}, ~\varepsilon > 0.
\end{equation}
Although it provides rather loose guarantees (for probability less than 0.001 and $\varepsilon=0.01$ it requires $M \approx 38000$, but if $\varepsilon=0.001$ $M \approx 3.8 \times 10^6$), it still might be helpful in some scenarios. The drawback of this approach is that it does not help to provide any guarantees for inverse empirical CDF.

Another way to estimate an error in Eq. \eqref{eq:hat_g} is to calculate the error from confidence intervals directly. As $h$ (smoothed--classifier prediction) is calculated via the Monte-Carlo sampling, we can declare its Normal-based confidence interval
\begin{equation}
\mathbb{P}\left(h \in(a, b)\right) \geqslant 1-\alpha \sim \frac{1}{\sqrt{M}}, 
\end{equation}
where $\alpha$ is a confidence value.
In probability space, the error decreases as 
\begin{equation}
\Delta_1 \sim \frac{C}{\sqrt{M}}(b-a)
\end{equation}
and provides domain-level error
\begin{equation}
\Delta_2 \sim\left|F^{-1}(1-b)-F^{-1}(1-a)\right|,
\end{equation}
that results in the following total perturbation 
\begin{equation}
\Delta_3 \sim \int_{z \in \mathbb{R}^n: \eta <  F^{-1}(1-a)} \eta(z, \hat{x}) \rho(z|\hat{x})dz - \int_{z \in \mathbb{R}^n: \eta>F^1(1-b)} \eta(z, \hat{x}) \rho(z|\hat{x})dz =
\\
\int_{z \in \mathbb{R}^n: \eta \in   \left[F^{-1}(1-b), F^{-1}(1-a) \right]} \eta(z, \hat{x}) \rho(z|\hat{x})dz
\end{equation}
With probability $1-\alpha$ interval $[\tilde g(h, \beta) - \Delta_3, \tilde g(h, \beta) + \Delta_3]$ covers the true value of $\tilde g(h, \beta)$,
 However, this might provide a useless estimation.

We also can apply the result for the convergence in distribution from \cite{Bartlett2013Inequalities}:
\begin{equation}
\sqrt{M}\left(F_M^{-1}(h)-F^{-1}(h)\right) \rightsquigarrow N\left(0, \frac{h(1-h)}{\rho^2\left(F^{-1}(h)\right)}\right) .
\end{equation}
This result is derived from the central limit theorem and has a logical non-linear dependency on $h$. It provides the uncertainty scale that standard deviation is decreasing as $\frac{1}{M}.$ Thus perturbation  $\tilde g(h, \beta)$ might simply bounded as

\begin{equation}
    \| \delta \tilde g(h, \beta) \| \leq  \sup_{z \in \mathbb{R}^n: \eta \in   \left[F^{-1}(1-h) - 3 \nu, F^{-1}(1-h) +3 \nu\right]} \| \eta \| 6 \nu, 
\end{equation}
where
\begin{equation}
    \nu = \frac{\sqrt{h(1-h)}}{\sqrt{M}\rho\left(F^{-1}(h)\right)}.
\end{equation}
Although we still do not know true $F^{-1}(h)$, we can apply this estimation for the confidence bounds of $h$ and $F_M^{-1}$. The results is still proportional to $\frac{1}{\sqrt{M}}$

\begin{equation}
    \|\delta g(\beta) \| = \sup_h \| \delta \tilde g(h, \beta) \|
\end{equation}

\begin{equation}
    \|\delta p(h) \| \approx \sup_{\beta}  \frac{\|\delta g(\beta) \| }{\sup_h \tilde g(h, \beta)} = \sup_{\beta}  \frac{\|\delta g(\beta) \| }{g(\beta)}.
\end{equation}
An error of analytical integration of $\xi$  is
\begin{equation}
    \delta_p  = \| \delta \int \frac{1}{p(h)} dh \| \sim \|\int \frac{- \|\delta p(h) \|}{p^2(h)} dh \| \sim \frac{C_3}{\sqrt{M}}
\end{equation}
An error in numerical integration is 
\begin{equation}
    \delta_{N_s} \leq \frac{1}{24 N_s^2}  \max_{h \in [0, 1]} \left| \frac{2 \left( p(h)^ {\prime}\right)^2 - p(h)^{\prime \prime} p(h)}{p^3(h)} \right|
\end{equation}
as we integrate on $[0, 1].$ Similar estimation might be obtained for $\hat g (\beta)$.

In addition, one can accelerate numerical convergence by applying Richardson extrapolation.

\section{Additional Discussion}
This section highlights some of the work's motivations and further research aspects.

When applied to the neural network, the GL framework allows the model to be certified to the specific transformation of the input data in the subspace of the parameter space by just computing the values of a specific function in the single point of the parameter space. Thus, non-ball vicinity certification for transformation parameters is achieved by construction (see Theorem 1).

Our approach is derived specifically for the classification problem. However, we want to mention that the method can be extended to the object detection problem (since the detection implies the classification problem for the corresponding bounding boxes). In practice, the method can also be applied to the image segmentation problem to certify the pixel-wise classification (note that in this case, the number of transformations would be limited, e.g., the method would not be applicable against image translation). The influence of segments’ connectivity on the certification guarantees is still an open question and could be considered as the future research direction.

Although our work focuses on the numerical certification of the neural networks rather than interpretability, one can try to apply interpretative methods, such as GradCAM and ScoreCAM. These methods improve the explainability of the neural networks, and their integration into the certification procedures is an interesting question for future research. However, adapting interpretative methods to the certification procedure raises a separate research question about aggregation over sampled transforms.

\end{document}